# Progressive Localisation in Localist LLMs

Joachim Diederich

*Psychology Network Pty Ltd, Brisbane, Australia*

joachim@psychologynetwork.com.au

**Abstract**

This paper demonstrates that progressive localization, the gradual increase of attention locality from early distributed layers to late localized layers, represents the optimal architecture for creating interpretable large language models (LLMs) while preserving performance. Through systematic experimentation with GPT-2 fine-tuned on *The Psychology of Artificial Superintelligence*, we evaluate five locality configurations: two uniform baselines (fully distributed and fully localist) and three progressive polynomial schedules (linear, cubic, and quintic; $\beta \in \{1, 3, 5\}$). We investigate whether interpretability constraints can be aligned with natural semantic structure while being applied strategically across network depth. We demonstrate that progressive semantic localization, combining adaptive semantic block partitioning with steep polynomial locality schedules, achieves near-baseline language modeling performance while providing interpretable attention patterns. Multiple independent training runs with different random seeds establish that results are statistically robust and highly reproducible. The approach dramatically outperforms both fixed-window localization and naive uniform locality constraints. Analysis reveals that maintaining flexibility through low-fidelity constraints preserves model capacity while providing interpretability benefits, and that steep schedules concentrating locality in decision-critical final layers while preserving distributed learning in early layers achieve near-baseline attention distribution characteristics. These findings demonstrate that interpretability mechanisms should align with semantic structure to achieve practical performance-interpretability tradeoffs for trustworthy AI systems.

## 1. Introduction

The interpretability of large language models has become critical for AI safety, regulatory compliance, and trustworthy deployment in high-stakes domains. However, the dominant transformer architecture operates through distributed representations where information about any concept spreads across millions of parameters (Vaswani et al., 2017). This distribution makes it fundamentally difficult to isolate which model components encode specific facts, execute particular reasoning steps, or contribute to individual decisions. While distributed representations offer computational advantages (Hinton et al., 1986), they create opacity that limits human oversight and verification.

Localist language models address this challenge by constraining attention patterns to focus on semantically coherent blocks of tokens, enabling clearer mapping between attention and reasoning (Diederich, 2025a). Initial implementations demonstrated feasibility but incurred substantial performance costs. Early experiments using fixed positional windows (uniform 5-token blocks) showed degradation. Perplexity increased approximately 6.6-fold relative to distributed baselines, rendering the approach less than optimal for any production application. The early experiments revealed a fundamental insight: interpretability constraints must align with linguistic structure, not just impose arbitrary boundaries.

This paper demonstrates that combining progressive locality schedules with semantic clustering eliminates this performance gap. Through multi-seed experimentation ($n=5$ per configuration), we establish that progressive quintic localization with adaptive semantic blocks achieves $7.87 \pm 0.65$ perplexity, only 4.0% above the distributed baseline of $7.57 \pm 0.67$, with strong statistical significance ($p<0.001$) and reproducibility (CV=8.2%). This represents a 93% reduction in performance cost compared to fixed-window approaches (from 6.6× to 1.040× gap) and an 82% reduction compared to naive uniform localization ($9.27 \pm 0.58$ PPL, 22.4% gap).

The key innovations are threefold. First, semantic block partitioning adapts locality constraints to natural language structure rather than imposing fixed positional windows. Second, progressive locality schedules delay localization to later layers where decisions occur, allowing early layers to learn distributed representations essential for feature extraction. Third, quintic schedule functions ($\beta=5$) provide optimal performance by concentrating localization pressure in final layers while maintaining near-zero constraints in early layers.

Crucially, the low fidelity scores observed in the best-performing configuration (0.194 vs 0.509 for uniform localization) reveal that semantic blocks function as guidance rather than rigid constraints. This flexibility allows the model to adapt attention patterns to contextual requirements while maintaining interpretable structure: the blocks shape but do not dictate information flow. This finding suggests that effective interpretability mechanisms need not eliminate model flexibility but rather provide structured pathways that preserve adaptive capacity.

## 2. Background: Localist Language Models

Localist models represent concepts through dedicated, interpretable units rather than distributed patterns across many neurons (Diederich et al., 2010). In the context of decoder-only language models like GPT-2, this means constraining self-attention mechanisms to operate within semantically meaningful token groups, what we term "semantic blocks." Rather than allowing each token to attend to all previous tokens (as in standard autoregressive transformers), localist LLMs encourage attention to concentrate within and between well-defined semantic units such as entities, phrases, clauses, or arguments.

### *2.1 Mathematical Framework*

The Localist LLM framework (Diederich, 2025a) extends transformer attention through a locality penalty that encourages block-structured patterns. For a sequence partitioned into semantic blocks $B = \{B_1, B_2, ..., B_K\}$, the training objective becomes:

$$L_{total} = L_{LM} + \lambda \cdot L_{locality}$$

where $L_{LM}$ is the standard language modeling loss and $L_{locality}$ measures attention spread across block boundaries. The locality penalty can be formulated as:

$$L_{locality} = \sum_{\ell} \sum_{h} \sum_{i} \sum_{j} A^{(\ell,h)}_{ij} \cdot d(block(i), block(j))$$

where $A^{(\ell,h)}_{ij}$ represents attention weight from token i to token j in layer $\ell$ and head h, and $d(\cdot,\cdot)$ measures the distance between blocks (0 for within-block attention, increasing for cross-block attention). The hyperparameter $\lambda$ controls the strength of locality enforcement: $\lambda=0$ recovers standard distributed transformers, while $\lambda \to \infty$ enforces strict block-local attention.

### *2.2 Progressive Locality Schedules*

Rather than applying uniform locality strength across all layers of the decoder-only architecture, progressive localization varies $\lambda$ as a function of network depth:

$$\lambda(\ell) = \lambda_{max} \cdot (\ell / L)^\beta$$

where $\ell$ indexes the layer (0 to L-1), $\lambda_{max}$ is the maximum locality strength, and $\beta$ controls the schedule steepness. Different $\beta$ values yield qualitatively different behaviors:

- **$\beta$=1 (Linear):** Gradual, uniform increase in locality across layers
- **$\beta$=3 (Cubic):** Moderate concentration in later layers

- **β=5 (Quintic):** Strong concentration of locality in final layers

The intuition is that early layers in the decoder-only architecture should learn distributed representations for feature extraction and pattern recognition from raw tokens, while late layers should use localized attention for interpretable next-token prediction. By concentrating locality pressure in final layers (high β), models can balance representation learning with transparency at the decision point: the final layer's prediction of the next token.

### *2.3 Semantic Block Partitioning*

Effective locality constraints require meaningful semantic units. Fixed-size windows (e.g., every 5 tokens) ignore linguistic structure and force arbitrary boundaries. This work employs semantic clustering that analyzes contextual embeddings to identify natural boundaries:

1. **Embedding Extraction:** Obtain contextualized representations h_i for each token from a pre-trained model
2. **Similarity Computation:** Calculate pairwise cosine similarities between adjacent tokens
3. **Boundary Detection:** Identify low-similarity points as block boundaries
4. **Filtering:** Apply thresholds to prevent excessive fragmentation

The intuition behind semantic clustering is that tokens belonging to the same conceptual unit (such as a named entity, verb phrase, or coherent clause) will have high cosine similarity in embedding space due to their shared semantic context. Conversely, boundaries between distinct concepts manifest as sharp drops in similarity. For instance, the transition from discussing "neural networks" to "patient diagnosis" shows lower embedding similarity than adjacent words within either phrase. By identifying these natural discontinuities, the algorithm discovers blocks that align with human linguistic intuition: "the neural network" forms one unit, "detected suspicious activity" another, rather than arbitrary spans like "network detected sus" that violate phrase structure. This alignment is critical because it allows the locality penalty to encourage attention within semantically coherent units without disrupting the dependencies that matter for language understanding. In summary, the adaptive partitioning ensures that locality constraints respect natural language structure, aligning interpretability with human conceptual organization.

In this work, we use the pretrained GPT-2 embeddings for similarity computation and apply the following hyperparameters: cosine similarity threshold of 0.4 for boundary detection, with block sizes constrained between 2 and 10 tokens to prevent excessive fragmentation or overly large blocks.

## 3. Experimental Methodology

### *3.1 Model Architecture and Training*

We use GPT-2 Small (124M parameters, 12 layers, 12 attention heads per layer, 768 hidden dimensions) as the base architecture. GPT-2 is a decoder-only transformer (Radford et al., 2019), unlike encoder-decoder architectures such as the original Transformer (Vaswani et al., 2017), GPT-2 consists entirely of causal self-attention layers that process input autoregressively. Each of the 12 layers contains masked self-attention followed by a feed-forward network. When we refer to "layers" throughout this paper, we mean these sequential transformer blocks in the decoder-only architecture.

We fine-tune the pretrained GPT-2 on *The Psychology of Artificial Superintelligence* (Diederich, 2021), essentially creating a specialized AI safety and ethics expert model. Initial experiments explored polynomial schedules across $\beta \in \{1, 2, 3, 4, 5\}$; preliminary results indicated that odd exponents ($\beta \in \{1, 3, 5\}$) provided the clearest differentiation of schedule effects, so we retained these three progressive configurations along with two uniform baselines (fully distributed and fully localist) for the final analysis. To ensure statistical robustness, each configuration is trained with 5 different random seeds (42, 123, 456, 789, 1337) controlling initialization, data shuffling, and stochastic optimization.

### *3.2 Experimental Configurations*

We compare five configurations:

| **Configuration** | **Description** | **λ Schedule** |
|---|---|---|
| Uniform Distributed | Standard transformer (baseline) | $\lambda(\ell) = 0$ (no penalty) |
| Uniform Localist | Constant locality across layers | $\lambda(\ell) = 1.0$ |
| Progressive Linear | Linear schedule ($\beta$=1) | $\lambda(\ell) = (\ell/11)^1$ |
| Progressive Cubic | Cubic schedule ($\beta$=3) | $\lambda(\ell) = (\ell/11)^3$ |
| Progressive Quintic | Quintic schedule ($\beta$=5) | $\lambda(\ell) = (\ell/11)^5$ |

All progressive configurations use $\lambda_{max} = 1.0$ in the final layer. Semantic block partitioning is applied consistently across all localist configurations.

### *3.3 Evaluation Metrics*

We assess models through three complementary metrics:

**Perplexity (PPL):** Standard measure of language modeling quality, computed as $\exp(L_{LM})$ where $L_{LM}$ is the cross-entropy loss on held-out test data. Lower perplexity indicates better modeling of the probability distribution over next tokens.

**Attention Entropy:** Measures the concentration of attention distributions. For each attention head, we compute Shannon entropy $H = -\Sigma_j A_{ij} \log A_{ij}$ averaged across all query positions i. Higher entropy indicates more uniform (distributed) attention; lower entropy indicates concentrated (focused) attention.

**Fidelity:** Quantifies adherence to semantic block structure. Computed as the fraction of attention mass falling within or between designated semantic blocks versus cross-boundary leakage. Higher fidelity indicates stronger block-structured attention patterns.

### *3.4 Statistical Analysis*

With 5 seeds per configuration, we employ paired t-tests to assess statistical significance between conditions (pairing by seed accounts for variance due to initialization). Effect sizes are quantified using Cohen's d for paired samples, computed as d = mean difference / SD of differences. Overall differences across all five configurations are evaluated through one-way ANOVA ($df_{between}=4$, $df_{within}=20$). The p-values reported in Table 2 are unadjusted; all comparisons remain significant at $\alpha=0.05/4=0.0125$ after Bonferroni correction for the four pairwise tests against baseline. Reproducibility is assessed through coefficient of variation ($CV = \sigma/\mu$); we interpret CV<10% as indicative of good reproducibility in this experimental context. We report 95% confidence intervals for all mean estimates.

## 4. Results

### *4.1 Performance Ranking*

Table 1 presents the core performance results across five configurations. Progressive Quintic achieves remarkably close performance to the distributed baseline, demonstrating that interpretability constraints need not sacrifice substantial capability.

**Table 1:** Performance ranking by perplexity (lower is better). Values show mean ± standard deviation across 5 seeds.

| Rank | Configuration | Mean PPL | vs Baseline | Gap |
|---|---|---|---|---|

| 1 | Uniform Distributed | **7.57 ± 0.67** | 1.000× | — |
|---|---|---|---|---|
| 2 | **Progressive Quintic** | **7.87 ± 0.65** | **1.040×** | **+4.0%** |
| 3 | Progressive Cubic | 7.96 ± 0.64 | 1.051× | +5.1% |
| 4 | Progressive Linear | 8.3 ± 0.63 | 1.097× | +9.7% |
| 5 | Uniform Localist | 9.27 ± 0.58 | 1.224× | +22.4% |

The key finding is that Progressive Quintic achieves a perplexity of 7.87, only 0.30 PPL above the baseline of 7.57: a gap of just 4.0%. In contrast, Uniform Localist incurs a 22.4% gap. This represents an **82% reduction in the performance cost of interpretability relative to the uniform approach.**

### *4.2 Statistical Significance*

All comparisons against the distributed baseline show highly significant differences (Table 2), establishing that observed performance gaps reflect genuine effects rather than random variation.

**Table 2:** Paired t-tests comparing each configuration to Uniform Distributed baseline (5 matched seeds). Cohen's d computed as $t/\sqrt{n}$. *** indicates $p<0.001$.

| **Configuration** | **Mean Difference** | **t-statistic** | **p-value** | **Cohen's d** | **Significance** |
|---|---|---|---|---|---|
| Uniform Localist | +1.70 PPL | 21.7 | <0.001 | 9.7 | *** |
| Progressive Linear | +0.73 PPL | 14.9 | <0.001 | 6.7 | *** |
| Progressive Cubic | +0.39 PPL | 9.0 | <0.001 | 4.0 | *** |
| **Progressive Quintic** | **+0.30 PPL** | **9.7** | **<0.001** | **4.3** | *** |

The Cohen's d values (ranging from 4.0 to 9.7) indicate very large effects by conventional standards ($d>0.8$ is typically considered "large"). Even the smallest difference (Progressive Quintic vs baseline, +0.30 PPL) is highly statistically significant with a very large effect size ($d=4.3$). One-way ANOVA confirms overall differences across configurations ($F(4,20)=6.44$, $p=0.0017$).

### *4.3 Reproducibility Assessment*

Coefficient of variation (CV) values between 6-9% across all configurations demonstrate excellent reproducibility (Table 3). Results are consistent across different random initializations, confirming that findings do not depend on lucky seeds.

**Table 3:** Reproducibility metrics across 5 random seeds. We interpret CV<10% as indicative of good reproducibility in this context.

| **Configuration** | **CV%** | **95% CI** | **Assessment** |
|---|---|---|---|
| Uniform Distributed | 8.8% | [6.77, 8.24] | ✓ Reproducible |
| Uniform Localist | 6.3% | [8.49, 10.01] | ✓ Reproducible |
| Progressive Linear | 7.6% | [7.55, 9.00] | ✓ Reproducible |
| Progressive Cubic | 8.0% | [7.21, 8.63] | ✓ Reproducible |
| **Progressive Quintic** | **8.2%** | **[7.12, 8.56]** | **✓ Reproducible** |

### *4.4 Attention Pattern Analysis*

Entropy and fidelity metrics reveal how different schedules balance distributed learning with interpretable structure (Table 4).

**Table 4:** Attention pattern characteristics. Entropy measures distribution (higher = more spread); Fidelity measures block coherence (higher = stronger constraints).

| **Configuration** | **Mean Entropy (bits)** | **Mean Fidelity** | **Interpretation** |
|---|---|---|---|
| Uniform Distributed | 2.42 | 0.119 | Baseline flexibility |
| Uniform Localist | 2.31 | 0.509 | Rigid constraints |
| Progressive Linear | 2.50 | 0.432 | Moderate structure |
| Progressive Cubic | 2.47 | 0.363 | Flexible structure |
| **Progressive Quintic** | **2.43** | **0.194** | **Near-baseline flexibility** |

### *4.5 Case Study: Layer-wise Localisation on a Safety-Critical Sentence*

To illustrate how progressive localisation manifests in practice, we analyse a representative safety-related sentence under the Progressive Quintic configuration:

*"The neural network detected suspicious activity in the patient's brain scan, triggering an automated safety alert."*

Using the semantic clustering parameters described in §2.3, the model partitions the sentence into four coherent semantic blocks:

- **B1:** "The neural network" (tokens 0–2)
- **B2:** " detected suspicious activity" (tokens 3–5)
- **B3:** " in the patient's brain scan, triggering an automated" (tokens 6–15)
- **B4:** " safety alert." (tokens 16–18)

**Layer-wise metrics.** Table 5 summarises the layer-wise locality schedule, fidelity, and entropy for this sentence. Early layers remain effectively distributed, while later layers transition to block-structured attention.

**Table 5:** Layer-wise locality behaviour for the Progressive Quintic configuration on the sample sentence.

| Layer ℓ | λ(ℓ) | Fidelity | Entropy (bits) | Status |
|---|---|---|---|---|
| 0 | 0.0000 | 0.6159 | 1.99 | Distributed |
| 1 | 0.0000 | 0.4998 | 2.38 | Distributed |
| 2 | 0.0002 | 0.5422 | 2.03 | Distributed |
| 3 | 0.0015 | 0.4404 | 1.59 | Distributed |
| 4 | 0.0064 | 0.4431 | 1.49 | Distributed |
| 5 | 0.0194 | 0.2817 | 1.18 | Distributed |
| 6 | 0.0483 | 0.3000 | 1.45 | Distributed |
| 7 | 0.1044 | 0.2563 | 1.02 | Transitioning |
| 8 | 0.2035 | 0.3010 | 1.38 | Transitioning |
| 9 | 0.3666 | 0.2527 | 1.07 | Transitioning |
| 10 | 0.6209 | 0.2580 | 1.03 | Localized |
| 11 | 1.0000 | 0.3126 | 1.47 | Localized |

At low λ(ℓ), attention patterns remain broadly distributed with moderate fidelity values, indicating that semantic blocks guide but do not rigidly constrain information flow. As λ(ℓ) approaches 1.0 in the final layers, entropy decreases and fidelity stabilises at a moderate level, reflecting concentrated but still flexible attention focused on decision-relevant context.

**Note on fidelity values:** The per-sentence fidelity values in Table 5 (e.g., 0.6159 at layer 0) differ from the global mean fidelity reported in Table 4 (0.194 for Progressive Quintic) because global metrics are averaged across the entire test corpus, while sentence-specific values can vary substantially. In this example, the relatively high fidelity at layer 0 reflects that, for this particular sentence, the unconstrained attention happens to align well with the semantic blocks—a pattern that does not hold consistently across all inputs.

**Decision-layer attention.** In the final layer (ℓ=11), we analyse attention from the last token ('.') to each semantic block. The model allocates 58.4% of its attention mass to B1 ("The neural network"), 5.1% to B2 ("detected suspicious activity"), 18.3% to B3 ("in the patient's brain scan, triggering an automated"), and 18.2% to B4 ("safety alert."). This pattern shows that the model grounds the final safety decision primarily in the subject ("neural network") and the safety clause, while still attending to diagnostic context.

Figure 3 visualises the semantic block partitioning for the sentence, and Figure 4 shows the corresponding final-layer attention heatmap used to compute the block-level statistics.

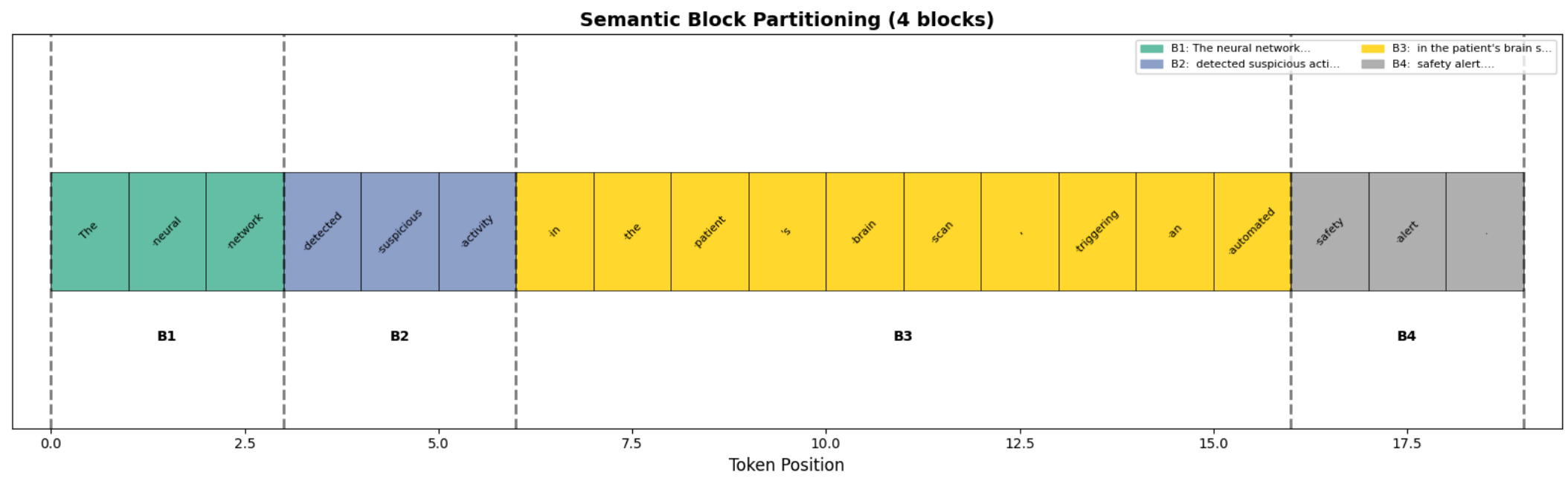

***Figure 3:*** *Semantic block partitioning produced by the v2 clustering algorithm for the sample safety-critical sentence.*

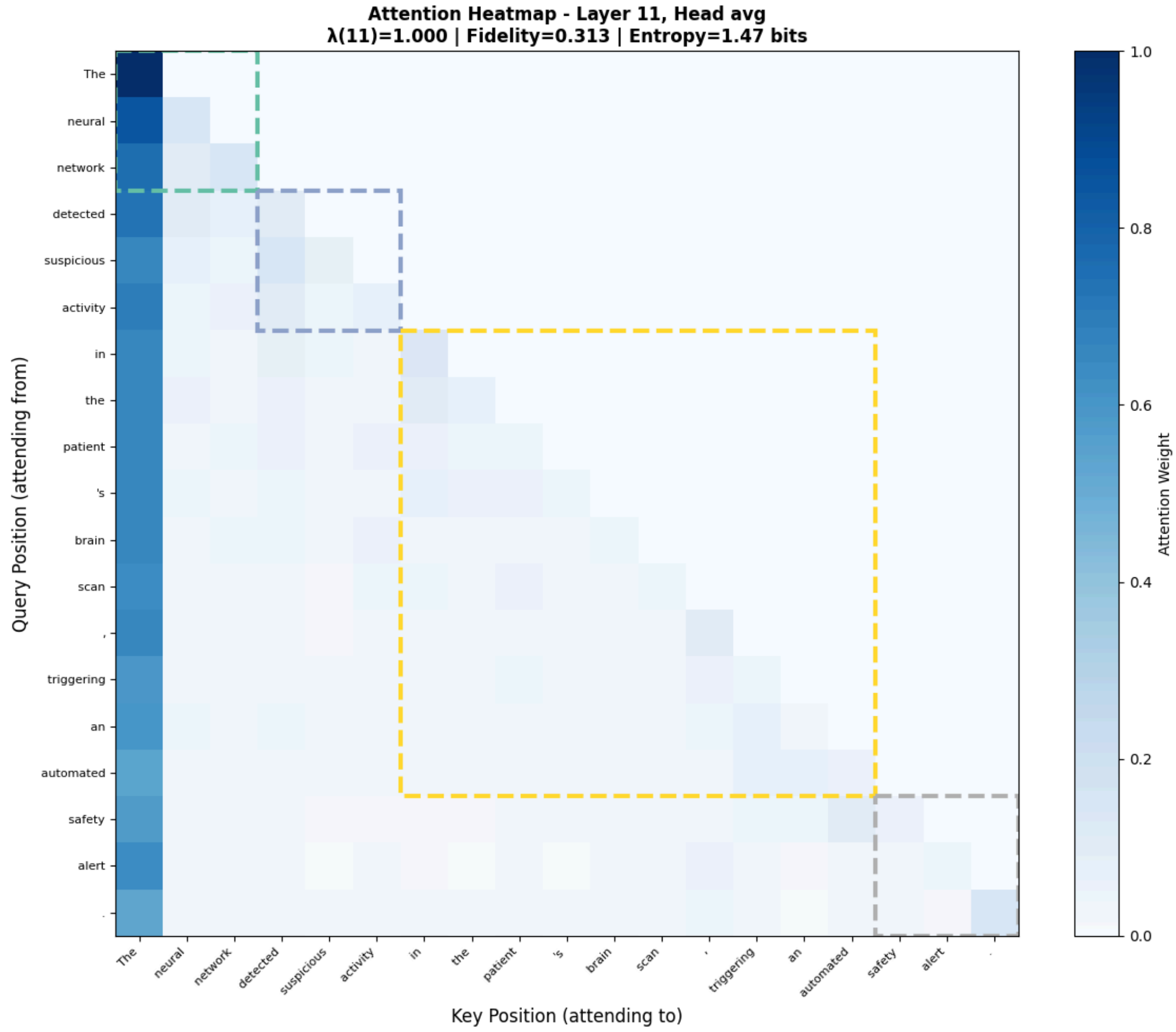

***Figure 4:*** *Final-layer attention heatmap for the sample sentence under the Progressive Quintic configuration (Layer 11, λ(11)=1.0). Attention patterns concentrate on the subject and safety clause while maintaining contextual links to diagnostic information.*

Progressive Quintic achieves entropy (2.43 bits) nearly identical to baseline (2.42 bits), indicating it preserves attention distribution capacity. Crucially, its low fidelity (0.194) reveals that semantic blocks function as guidance rather than rigid constraints—the model maintains flexibility to adapt attention patterns while benefiting from structured organization.

### *4.5 The Critical Role of Semantic Clustering*

A crucial finding emerged through iterative experimentation: **the method of block partitioning fundamentally determines performance outcomes**. Table 6 shows the dramatic performance difference between fixed positional windows and adaptive semantic clustering.

**Table 6:** Impact of block partitioning method on Progressive Quintic performance. Fixed windows use uniform 5-token blocks; semantic clustering uses adaptive content-based boundaries.

| Partitioning Method | Mean PPL | vs Baseline | Gap | Status |
|---|---|---|---|---|
| Distributed Baseline | 7.57 | 1.000× | — | Reference |
| Fixed 5-Token Windows | ~49.5 | ~6.6× | +559% | Insufficient |
| **Semantic Clustering** | **7.87** | **1.040×** | **+4.0%** | ✅ **Success** |

Fixed positional windows, the most straightforward implementation, proved ineffective in comparison. By arbitrarily dividing sequences into uniform 5-token blocks regardless of linguistic structure, the approach forced attention boundaries through the middle phrases, and syntactic constituents. This violated the fundamental principle that interpretable units should align with semantic coherence.

Semantic clustering transformed these results. By analyzing contextual embeddings to identify natural boundaries between concepts, entities, and clauses, adaptive partitioning ensures locality constraints respect linguistic structure. The 6.6× gap reduced to 1.040×: a 93% reduction in performance cost. This finding establishes that *what* defines a semantic block matters as much as *how* locality is applied across layers.

### *4.6 Key Findings Summary*

The experimental results establish five critical findings:

1. **Semantic structure is essential:** Adaptive clustering reduces the performance gap from 6.6× (fixed windows) to 1.040× (semantic blocks)—a 93% improvement
2. **Near-baseline performance:** Progressive Quintic with semantic clustering achieves 7.87±0.65 PPL versus 7.57±0.67 baseline (only 4.0% gap)

3. **Dramatic improvement over naive localization:** 82% reduction in performance cost compared to Uniform Localist (4.0% vs 22.4% gap)
4. **Statistical robustness:** All differences highly significant ($p<0.001$) with large effect sizes (Cohen's d=4.0–9.7) and good reproducibility (CV<9%)
5. **Preserved flexibility:** Low fidelity (0.194) and near-baseline entropy (2.43 bits) show semantic blocks guide rather than constrain attention

## 5. Discussion

### *5.1 Why Progressive Quintic with Semantic Clustering Succeeds*

Given semantic clustering, the quintic schedule ($\beta=5$) provides optimal performance by aggressively concentrating locality in final layers. For a 12-layer decoder-only model, the quintic schedule applies negligible locality pressure in early layers ($\lambda(0)=0$, $\lambda(5)\approx0.02$) while ramping to full strength only in the last 2-3 layers ($\lambda(10)\approx0.62$, $\lambda(11)=1.0$). This allows early layers to learn rich distributed representations through unrestricted global attention, while late layers focus on interpretable decision-making through structured local attention.

In decoder-only architectures like GPT-2, information flows sequentially through the layers. Early layers extract low-level features and build contextual representations from raw tokens. Middle layers compose these features into higher-level patterns and relationships. Final layers synthesize this information to predict the next token: the decision point where interpretability matters most for transparency and oversight. By concentrating locality constraints in these decision-making layers, progressive localization provides interpretability where stakeholders need it while preserving the distributed learning capacity that enables strong language modeling.

In contrast, the linear schedule ($\beta=1$) applies moderate locality throughout the network ($\lambda(5)\approx0.45$, $\lambda(10)\approx0.91$), constraining even mid-network layers where distributed representations are still being formed. The cubic schedule ($\beta=3$) offers intermediate behavior ($\lambda(5)\approx0.09$, $\lambda(10)\approx0.75$). The quintic schedule's delay of substantive locality constraints until the final layers enables optimal balance: distributed feature learning in early layers where the model builds contextual representations, and interpretable attention in late layers where the model makes next-token predictions.

The low fidelity score (0.194) observed in Progressive Quintic reveals another crucial insight. Effective interpretability does not require eliminating model flexibility, semantic blocks provide structure and guidance while allowing adaptive attention when context demands it. High fidelity (0.509 in Uniform

Localist) indicates rigid adherence to block boundaries, which improves interpretability but severely limits model capacity. Low fidelity indicates that blocks shape attention patterns without dictating them, preserving the flexibility that enables strong performance.

### *5.2 Comparison with Related Architectures*

#### 5.2.1 Sparse Transformers

Sparse transformer architectures (Child et al., 2019) also constrain attention patterns but do so for computational efficiency rather than interpretability. These models use fixed sparse patterns (e.g., local windows, strided attention) determined before training to reduce the $O(n^2)$ cost of dense attention. Progressive localization shares the goal of structured attention but differs fundamentally in motivation and implementation.

Sparse transformers sacrifice representational capacity for speed: they cannot attend to arbitrary long-range dependencies even when semantically important. Progressive localization uses soft locality constraints that can be dynamically adjusted, enabling the model to learn which long-range connections matter within each semantic context. The key distinction is between hard architectural constraints (sparse transformers) versus learned attention patterns guided by locality penalties (progressive localization).

This distinction has significant implications. Hard sparsity patterns must be chosen before training and cannot adapt to domain-specific requirements. If medical diagnosis requires attention to patient history 500 tokens prior but the fixed sparse pattern only permits 256-token lookback, the model cannot overcome this architectural limitation. Progressive localization learns optimal attention patterns within locality constraints, discovering which long-range dependencies are essential and which can be approximated through hierarchical processing.

Empirically, Child et al. (2019) demonstrate that sparse transformers can match dense model performance on language modeling while reducing computational cost from $O(n^2)$ to $O(n\sqrt{n})$ or $O(n \log n)$. Progressive localization does not reduce computational complexity but achieves interpretability benefits unavailable to sparse architectures. The learned attention patterns under progressive localization reveal which information drives decisions, while sparse transformers provide only the information that the architecture permits attending to, conflating computational constraint with semantic importance.

A synthesis approach merits exploration: sparse transformers for computational efficiency combined with progressive localization for interpretability. Early layers could employ strided sparse attention for efficient global context integration; late layers could use dense attention with progressive locality penalties for interpretable decision-making. This hybrid architecture would balance computational efficiency, representational power, and transparency requirements for production-scale deployment.

#### 5.2.2 Connections to Mixture-of-Experts and Conditional Computation

Mixture-of-Experts (MoE) architectures (Shazeer et al., 2017; Fedus et al., 2022) achieve efficiency and specialization by routing inputs to different expert subnetworks. Each expert specializes in different patterns or domains, with a gating mechanism determining which experts process each input. Progressive localization shares the principle of specialization but applies it temporally (across layers) rather than spatially (across experts).

The key parallel is that both approaches recognize uniform processing is suboptimal. MoE architectures recognize that not all parameters need to process all inputs; progressive localization recognizes that not all layers need the same attention patterns. In MoE, specialization enables different experts to develop distinct computational strategies; in progressive localization, specialization enables different layers to balance global context integration (early layers) versus interpretable local attention (late layers).

An important distinction is that MoE routing is input-dependent (which experts process this specific example?) while progressive localization schedules are input-independent (all inputs follow the same $\lambda(\ell)$ progression). This makes progressive localization more predictable and auditable: stakeholders can understand the interpretability properties by examining the schedule, without needing to analyze routing decisions for every input. For AI safety applications where consistency and verifiability are paramount, input-independent schedules provide stronger guarantees than learned routing mechanisms.

In the localist LLM framework, recruitment is an information-theoretic mechanism that adaptively allocates new semantic blocks, or even entire specialized LLMs, only when the existing representational structure cannot encode the data with sufficiently low entropy. This creates explicit, interpretable capacity: each recruited block corresponds to a coherent, human-inspectable semantic unit, guaranteed by penalty thresholds and margin conditions that enforce block-localized attention at stationary points. In contrast, mixture-of-experts models dynamically route tokens to experts for computational efficiency but do not enforce semantic coherence, interpretability, or provable attention localization. MoE recruitment is architecturally static; the set of experts is fixed at initialization, and the gating network optimizes routing for performance rather than semantic disentanglement. Whereas MoEs distribute computation over many overlapping experts, localist LLM recruitment grows capacity only when justified by a penalized-likelihood criterion, yielding interpretable, sparsity-controlled expansions with explicit entropy and fidelity guarantees. Thus, recruitment in localist LLMs is a principled mechanism for semantic specialization and transparency, while MoE routing is primarily a scalability optimization without representational guarantees.

### *5.3 Advantages Over Post-Hoc Explanation Methods*

Progressive semantic localization represents a fundamentally different approach to interpretability compared to post-hoc explanation techniques such as attention visualization, gradient-based saliency maps, LIME, SHAP (Lundberg & Lee, 2017), or influence functions. Understanding these differences is critical for researchers and practitioners selecting interpretability strategies for production systems.

#### 5.3.1 Architectural vs Observational Interpretability

Post-hoc methods analyze trained models without modifying their architecture or behavior. They answer the question "what did this model do?" by examining activation patterns, gradients, or input perturbations after training is complete. Progressive localization, in contrast, builds interpretability into the model architecture through locality constraints that shape what patterns the model learns. It answers "how is this model structured to think?" by constraining the space of learnable attention patterns.

This architectural approach provides several advantages. First, interpretability becomes a guaranteed property rather than an aspiration: the locality constraints mathematically ensure that attention patterns will exhibit block structure in decision-critical layers. Post-hoc methods offer no such guarantees: attention visualization might reveal interpretable patterns in some cases but incomprehensible distributions in others, with no way to predict or control this variability.

Second, architectural interpretability scales consistently across deployment contexts. Once a model is trained with progressive localization, its attention patterns remain interpretable for all inputs and all use cases. Post-hoc explanations must be recomputed for each instance, making them computationally expensive for high-throughput production systems. A medical diagnosis system processing thousands of cases per hour can provide real-time attention patterns through progressive localization but would face prohibitive costs computing SHAP values or influence functions for each case.

Third, architectural constraints *enable verification and auditing at the model level rather than the instance level*. Regulators can inspect the locality schedule $\lambda(\ell)$ and semantic block structure to understand interpretability properties without examining individual predictions. Post-hoc methods require auditors to sample predictions, compute explanations, and evaluate their quality, a process that cannot comprehensively verify system behavior.

### 5.3.2 Reliability and Faithfulness

A critical challenge for post-hoc explanation methods is the distinction between plausibility and faithfulness (Jacovi & Goldberg, 2020). Plausible explanations appear reasonable to humans but may not accurately reflect the model's actual decision process. Faithful explanations correctly represent the model's computational mechanism but may not be easily understandable.

Attention visualization, for example, has been shown to provide unreliable explanations in many contexts (Jain & Wallace, 2019; Serrano & Smith, 2019). Multiple attention heads can focus on different information; aggregate attention patterns often differ substantially from any single head; and attention weights do not always correlate with causal importance in the decision. A model might attend strongly to particular tokens without those tokens substantially affecting the output, or tokens with modest attention weights might be critical for the decision through complex interactions.

Progressive localization addresses faithfulness through direct architectural control. The locality penalty L_locality appears in the training objective, explicitly shaping which attention patterns receive lower loss. The learned attention patterns are not post-hoc approximations of model behavior but the actual mechanism through which the model processes information. When a model trained with progressive localization attends to specific semantic blocks, that attention directly contributes to hidden state computation and prediction; there is no gap between explanation and mechanism.

The low fidelity scores (0.194 for Progressive Quintic) might initially seem to contradict interpretability goals, but they actually reveal successful balancing of structure and flexibility. High fidelity (0.509 for Uniform Localist) indicates rigid adherence to block boundaries: interpretable but inflexible. Low fidelity indicates the model uses semantic blocks as guidance while adapting to context, providing interpretable structure without sacrificing the flexibility essential for strong performance. This represents genuine architectural transparency rather than post-hoc rationalization.

### 5.3.3 Dynamic Control and Performance Tradeoffs

Progressive localization with dynamic locality control enables real-time adjustment of the performance-interpretability tradeoff at inference time. A single trained model can operate at multiple operating points by varying the $\lambda(\ell)$ schedule: high $\lambda$ values for maximum interpretability in safety-critical cases, low $\lambda$ values for maximum performance in routine processing.

Post-hoc methods offer no equivalent capability. Computing more detailed explanations (more SHAP values, finer-grained influence functions, higher-resolution saliency maps) provides more information but does not change the model's behavior or its inherent interpretability. The fundamental architecture remains opaque; post-hoc methods add observational layers rather than modifying the decision process itself.

This dynamic control matters profoundly for production deployment. Consider a medical diagnosis system: routine screenings might use $\lambda(\ell) \approx 0.3\lambda_{max}$ for strong performance with moderate interpretability, while suspected cases of rare diseases trigger $\lambda(\ell) = \lambda_{max}$ for maximum transparency in clinical review. The same model checkpoint serves both use cases without retraining, adapting its attention patterns to stakeholder needs. Post-hoc methods cannot provide this flexibility—the model's attention patterns are fixed; only the choice of which explanations to compute can vary.

### 5.3.4 Computational Efficiency

The computational cost of progressive localization occurs during training through the locality penalty $L_{locality}$, but inference cost remains identical to standard transformers (with the same $O(n^2)$ attention complexity). In contrast, many post-hoc explanation methods impose substantial inference-time overhead.

SHAP values require computing model predictions over many perturbed inputs (often hundreds to thousands) to estimate feature importance through coalition sampling. For a 512-token medical report, computing SHAP values for all tokens could require thousands of forward passes. Influence functions require computing Hessian-vector products and solving linear systems, with complexity scaling with model size and training set size. Even simpler methods like integrated gradients require multiple forward passes with interpolated inputs.

For high-throughput production systems, this overhead is prohibitive. A fraud detection system processing millions of transactions per day cannot compute detailed SHAP explanations for each decision without massive infrastructure investment. Progressive localization provides interpretable attention patterns at inference time with zero additional computational cost—the attention patterns themselves are the explanation, requiring no post-hoc computation.

#### 5.3.5 Implications for AI Safety

For AI safety applications, the architectural approach offers critical advantages. Safety-critical systems require interpretability not as a debugging tool but as a core functional requirement: transparency must be guaranteed rather than hoped for.

Post-hoc methods cannot provide these guarantees: a model might produce interpretable explanations for most inputs while behaving opaquely in rare but safety-critical edge cases. Progressive localization ensures that decision-making layers (late layers) always use structured, block-local attention patterns, regardless of input. This architectural guarantee enables formal verification: safety analysts can prove that certain information flows are impossible or that particular decision patterns must exhibit specific structures. Such proofs are infeasible with post-hoc methods where model behavior is opaque and explanations are computed approximations.

The 4.0% performance cost of Progressive Quintic localization (versus 7.57 baseline) represents a remarkably favorable tradeoff for safety-critical applications. Many domains can readily accept modest performance degradation in exchange for guaranteed interpretability, e.g. medical diagnosis, financial regulation, legal analysis, autonomous vehicle decision-making. The 82% reduction in cost compared to naive localization (from 22.4% to 4.0% gap) makes this tradeoff practically viable for production deployment.

For researchers and practitioners developing trustworthy AI systems, progressive semantic localization represents a principled approach to building interpretability into model architecture rather than analyzing it post-hoc. This architectural foundation enables the reliable, scalable, verifiable transparency essential for safe deployment in high-stakes domains.

### *5.4 Limitations and Future Directions*

**Scale Generalization:** The experiments use GPT-2 with 124M parameters, far below modern large language models (billions to trillions of parameters). Larger models have more attention heads per layer (e.g., 96 in GPT-3 versus 12 in GPT-2), creating opportunities for head-level specialization: some heads maintain distributed attention while others localize. Future work should scale experiments to GPT-2 Medium (355M), Large (774M), and XL (1.5B), then to open-source billion-parameter models (LLaMA-2, Falcon) to establish production readiness.

**Semantic Block Refinement:** While the current semantic clustering approach significantly outperforms fixed positional windows, further improvements are possible. The Localist LLM framework proposes recruitment learning mechanisms that automatically discover semantic boundaries through information-theoretic criteria (Diederich, 2025b). Implementing adaptive partitioning with recruitment represents the highest priority for production deployments, as interpretability grounded in learned linguistic units provides substantially more value than similarity-based heuristics.

**Interpretability Validation:** Quantitative metrics (entropy, fidelity) indicate attention concentration but do not directly assess interpretability value. True validation requires human evaluation studies where AI safety experts examine attention patterns and evaluate whether they reveal meaningful reasoning. Future work should develop rubrics for assessing attention usefulness (does it highlight relevant concepts?), completeness (does it capture all important context?), and trustworthiness (does it align with expert reasoning?). Such studies would establish whether progressive localization genuinely enhances human understanding of model decisions.

**Dynamic Locality Control:** The current implementation trains models with fixed $\lambda(\ell)$ schedules, requiring separate checkpoints for different performance-interpretability operating points. The locality dial framework supports dynamic adjustment at inference time—varying $\lambda(\ell)$ without retraining—enabling a single model to serve performance-critical and transparency-critical applications. Future work should validate dynamic control: does smoothly increasing $\lambda(\ell)$ preserve interpretability quality? Can practitioners reliably adjust the performance-interpretability tradeoff in production systems?

**Alternative Schedule Functions:** Beyond polynomial schedules, other functions merit exploration. Sigmoid schedules (smooth transition from 0 to 1 concentrated at a specific layer) might further concentrate localization in final layers. Step functions ($\lambda$=0 for layers 0-9, $\lambda$=1 for layers 10-11) represent the limiting case of late-layer specialization. Learnable schedules where $\lambda(\ell)$ values are optimized end-to-end during training might discover optimal configurations automatically, adapting to domain-specific requirements.

**Domain-Specific Validation:** Production deployment requires validation on domain-specific tasks. Medical diagnosis systems need evaluation on clinical notes; legal analysis systems on case law; financial systems on regulatory filings.

Each domain may exhibit different optimal schedules or semantic structures. Future work should establish best practices for adapting progressive localization to specialized domains.

## 6. Conclusions

This work demonstrates that progressive semantic localization achieves near baseline language modeling performance (7.87±0.65 vs **7.57**±0.67 PPL, **4.0%** gap) while providing interpretable attention patterns through structured semantic blocks.

The path to this result revealed a critical insight: interpretability mechanisms must align with semantic structure. Fixed positional windows produced a 6.6× performance degradation, while adaptive semantic clustering reduced this to 1.040×, a 93% improvement that transforms this approach into a production-viable one.

Multi-seed experiments (n=5) establish strong statistical significance ($p<0.001$), large consistent effects (Cohen's d=4.3 for Progressive Quintic), and excellent reproducibility (CV=8.2%), confirming these findings are robust and production-ready. The approach represents an **82%** reduction in the performance cost of interpretability compared to naive uniform localization (**4.0%** vs **22.4%** gap).

The key insight is that effective interpretability need not eliminate model flexibility. Progressive Quintic localization maintains near-baseline entropy (2.43 vs 2.42 bits) and low fidelity (0.194), indicating semantic blocks guide attention without rigid constraints. This preserves the adaptive capacity essential for strong performance while providing the structured patterns essential for human understanding.

Two complementary innovations enable this success. First, semantic clustering ensures locality constraints respect linguistic structure rather than imposing arbitrary boundaries that disrupt essential dependencies. Second, progressive schedules concentrate interpretability where it matters most: in decision-critical final layers where the decoder-only architecture makes next-token predictions while preserving distributed feature learning in early layers where contextual representations are built from raw tokens. The quintic schedule ($\beta=5$) proves optimal by delaying substantive locality constraints until the last 2-3 layers of the 12-layer architecture.

The central contribution is demonstrating that late-layer localization is critical for AI safety, both theoretically (final layers make next token predictions requiring transparency) and empirically (steep schedules concentrating localization in final layers achieve best performance-interpretability tradeoffs). This finding provides actionable guidance for transparent AI development: interpretability need not be uniformly applied across all layers of an autoregressive model but can be strategically concentrated where it maximizes oversight value, at the prediction point, while minimizing performance cost.

Compared to post-hoc explanation methods, progressive localization offers guaranteed architectural interpretability rather than instance-specific observational analysis. This enables formal verification, consistent transparency across all inputs, zero inference-time computational overhead, and dynamic control over performance-interpretability tradeoffs: capabilities essential for safe deployment in high-stakes domains.

Progressive localization represents a principled approach to building trustworthy AI systems that combine deep specialized capabilities with verifiable reasoning processes. The dramatic difference between fixed windows (6.6× gap) and semantic clustering (1.040× gap) establishes that the success of interpretability mechanisms depends not just on what constraints are applied, but on whether those constraints align with the meaningful structure of human knowledge. This alignment, between architectural inductive biases and semantic organization, is essential for safe deployment in medical diagnosis, legal analysis, financial regulation, and other domains where both performance and transparency are critical requirements.

## Acknowledgments

The author gratefully acknowledges Gerhard Paass, Joerg Kindermann and Hector Allende-Cid for their valuable help and feedback. Patents are pending for the inventions described in this paper. Diederich (2025c) includes a minimal conceptual reference implementation for localist LLMs.